\documentclass{article}

\usepackage[preprint]{nips_2018}

\usepackage[utf8]{inputenc} 
\usepackage[T1]{fontenc}    
\usepackage{hyperref}       
\usepackage{url}            
\usepackage{booktabs}       
\usepackage{amsfonts}       
\usepackage{nicefrac}       
\usepackage{microtype}      
\usepackage{wrapfig}
\usepackage{subcaption}
\usepackage{multicol}

\usepackage{mathtools}
\usepackage{color}
\usepackage[ruled]{algorithm2e}

\newcommand{\Bernpm}{\ensuremath{\textrm{Bern}_{\pm}}}
\usepackage{amssymb}
\newcommand{\mathbold}[1]{\ensuremath{\boldsymbol{\mathbf{#1}}}}

\newcommand{\mba}{\mathbold{a}}

\newcommand{\mbh}{\mathbold{h}}

\newcommand{\mbm}{\mathbold{m}}
\newcommand{\mbn}{\mathbold{n}}

\newcommand{\mbp}{\mathbold{p}}

\newcommand{\mbs}{\mathbold{s}}

\newcommand{\mbu}{\mathbold{u}}
\newcommand{\mbv}{\mathbold{v}}

\newcommand{\mbx}{\mathbold{x}}

\newcommand{\mbz}{\mathbold{z}}

\newcommand{\mbB}{\mathbold{B}}

\newcommand{\mbI}{\mathbold{I}}

\newcommand{\mbW}{\mathbold{W}}
\newcommand{\mbX}{\mathbold{X}}
\newcommand{\mbY}{\mathbold{Y}}

\newcommand{\mbepsilon}{\mathbold{\epsilon}}

\newcommand{\mbiota}{\mathbold{\iota}}

\newcommand{\mbmu}{\mathbold{\mu}}

\newcommand{\mbrho}{\mathbold{\rho}}
\newcommand{\mbsigma}{\mathbold{\sigma}}

\newcommand{\mbtheta}{\mathbold{\theta}}

\title{Probabilistic Binary Neural Networks}

\author{
  Jorn W.T. Peters \\
  DELTA Lab, University of Amsterdam \\
  Amsterdam, The Netherlands \\
  \texttt{j.w.t.peters@uva.nl}
  \And
  Max Welling\\
  DELTA Lab, University of Amsterdam \\
  Amsterdam, The Netherlands \\
  \texttt{m.welling@uva.nl}
}

\begin{document}

\maketitle

\begin{abstract}
Low bit-width weights and activations are an effective way of combating the
increasing need for both memory and compute power of Deep Neural
Networks. In this work, we present a probabilistic training method for
Neural Network with both binary weights and activations, called
BLRNet. By embracing stochasticity during training, we circumvent the
need to approximate the gradient of non-differentiable functions such
as $\textrm{sign}(\cdot)$, while still obtaining a fully Binary Neural
Network at test time. Moreover, it allows for anytime ensemble
predictions for improved performance and uncertainty estimates by
sampling from the weight distribution. Since all operations in a
layer of the BLRNet operate on random variables, we introduce
stochastic versions of Batch Normalization and max pooling, which
transfer well to a deterministic network at test time.  We evaluate
the BLRNet on multiple standardized benchmarks.
  


\end{abstract}

\section{Introduction}
Deep Neural Networks are notorious for having vast memory and
computation requirements, both during training and test/prediction
time. As such, Deep Neural Networks may be unfeasible in various
environments such as on-body devices (such as hearing aids) due to heat
dissipation, battery driven devices due to power requirements,
embedded systems because of memory requirements, or real-time system
in which constraints are imposed by a limited economical budget.
Hence, there is a clear need for Neural Networks that can operate in
resource limited environments.

One method for reducing the memory and computational requirements for
Neural Networks is to reduce the bit-width of the parameters and
activations of the Neural Network. This can be achieved either during
training (e.g., \cite{ullrich2017soft, achterhold2018variational}) or
using post-training mechanisms (e.g., \cite{louizos2017bayesian},
\cite{han2015deep}). By taking the reduction of the bit-width for
weights and activations to the extreme, i.e., a single bit, one
obtains a Binary Neural Network. Binary Neural Networks have several
advantageous properties, i.e., a $32\times$ reduction in memory
requirements and the forward pass can be implemented using XNOR
operations and bit-counting, which results in a $58\times$
speedup~\citep{rastegari2016xnor}. Moreover, Binary Neural Networks are
more robust to adversarial examples~\citep{galloway2018attacking}.

\cite{shayar2017learning} introduced a probabilistic training method
for Neural Networks with binary weights, but allow for full precision
activations. In this paper, we propose a probabilistic training method
for Neural Networks with both binary weights and binary activations,
which are even more memory and computation efficient.  In short, we
train a stochastic Binary Neural Network by leveraging both the local
reparametrization trick~\citep{kingma2015variational} and the Concrete
distribution~\citep{maddison2016concrete,jang2016categorical}. At test
time, we obtain a single deterministic Binary Neural Network or an
ensemble of Binary Neural Networks by sampling from the parameter
distribution. An advantage of our method is that we can take samples
from the parameter distribution indefinitely---without
retraining. Hence, this method allows for anytime ensemble predictions
and uncertainty estimates. The stochastic network has a clear Bayesian
interpretation: the parameter distribution $p(\mbW)$ of the stochastic
network is a variational approximation to the true posterior
$p(\mbW|\mbX, \mbY)$, where $(\mbX, \mbY)$ denote the data, assuming a
uniform prior on the weights. This interpretation may be used for
further pruning of the network or allows for the introduction of more
sophisticated priors.  Note that while in this work we only consider
the binary case, our method supports any discrete distribution over
weights and activations.

In the proposed method, binary activations are sampled as the
very last operation in each layer.  As such, any other operation that
is normally applied to the pre-activation must be applied to random
variables. One of the contributions of this paper is the definition
of batch-normalization and max-pooling for random variables.
Our experiments show that these operations transfer well to a
non-stochastic operation in a deterministic network -- after
re-estimation of the batch norm statistics.


\section{Binary Neural Networks}
Binary and low precision neural networks have received significant
interest in recent years. Most similar to our work, in terms of the
final neural network, is the work on Binarized Neural Networks
by~\cite{hubara2016binarized}. in this work a real-valued shadow weight is
used and binary weights are obtained by binarizing the shadow
weights. Similarly the pre-activations are binarized using the same
binarization function. In order to back-propagate through the
binarization operation the straight-through
estimator~\citep{hintonst} is used. Several extensions to Binarized Neural
Networks have been proposed which --- more or less --- qualify as binary
neural networks: XNOR-net~\citep{rastegari2016xnor} in which
the real-valued parameter tensor and activation tensor is approximated
by a binary tensor and a scaling factor per
channel. ABC-nets~\cite{lin2017towards} take this approach one step
further and approximate the weight tensor by a linear combination of
binary tensors. Both of these approaches perform the linear operations
in the forward pass using binary weights and/or binary activations,
followed by a scaling or linear combination of the
pre-activations. In~\cite{mcdonnel2018training}, similar methods
to~\cite{hubara2016binarized} are used to binarize a wide
resnet~\citep{zagoruyko2016wide} to obtain results on ImageNet very
close to the full precision performance.  Another method for training
binary neural networks is Expectation
Backpropagation~\citep{soudry2014expectation} in which the central
limit theorem and online expectation propagation is used to find an
approximate posterior. This method is similar in spirit to ours, but
the training method is completely different.

\subsection{Binary Weight Network using Local Reparameterization}
In this section we describe the binary local reparametrization method
by~\cite{shayar2017learning} for a single layer in a neural
network. Assume a layer with $K\times K$ dimensional stochastic binary
weights $[\mbB_{ij}] \sim p(\mbB)$, such that
\begin{align}
  p(\mbB_{ij} = -1) = \sigma(\mbW_{ij}),\ \text{and}\ 
  p(\mbB_{ij} = +1) = 1 - p(\mbB_{ij} = -1).
\end{align}
Since $\mbB$ is a random variable, $\mbz = \mbB\mbh$ is also a random
variable, where $\mbh$ is the activation of the previous layer. From
the (Lyapunov) Central Limit Theorem (CLT), it follows that $\mbz$ is normally
distributed, specifically:
\begin{align}
  \mbz_i \sim \mathcal{N}(\mbmu_i, \mbsigma^2_i) =
  \mathcal{N}(\sum_{j=1}^K \mbh_j \mathbb{E}[\mbB_{ij}], \sum_{j=1}^K \mbh_j^2 \mathbb{V}[\mbB_{ij}]).
\end{align}
Hence, we obtain a distribution over pre-activations. From this
distribution we can easily sample using the reparameterization
trick~\citep{kingma2013auto} to obtain a real-valued pre-activation,
i.e.,
\begin{align}
  \mba_i = \mbmu_i + \mbsigma_i \odot \mbepsilon, \quad \text{where $\mbepsilon \sim \mathcal{N}(0, 1)$.}
\end{align}
The combination of the CLT approximation and sampling using the
reparameterization trick is also known as the local reparameterization
trick~\citep{kingma2015variational}.  Given the sample $\mba_i$ we can
proceed as usual and apply batch normalization, max-pooling and
non-linearities. At test time, instead of using the local
reparameterization trick, a binary weight matrix $\hat{\mbB} \sim
p(\mbB)$ is sampled and used for all test data, i.e., $\mba =
\hat{\mbB}\mbh$.
\pagebreak
\section{Binary Local Reparameterization Network}
\label{sec:blrnet}
\begin{wrapfigure}{r}{8cm}
  \centering
  \includegraphics[width=0.6\textwidth]{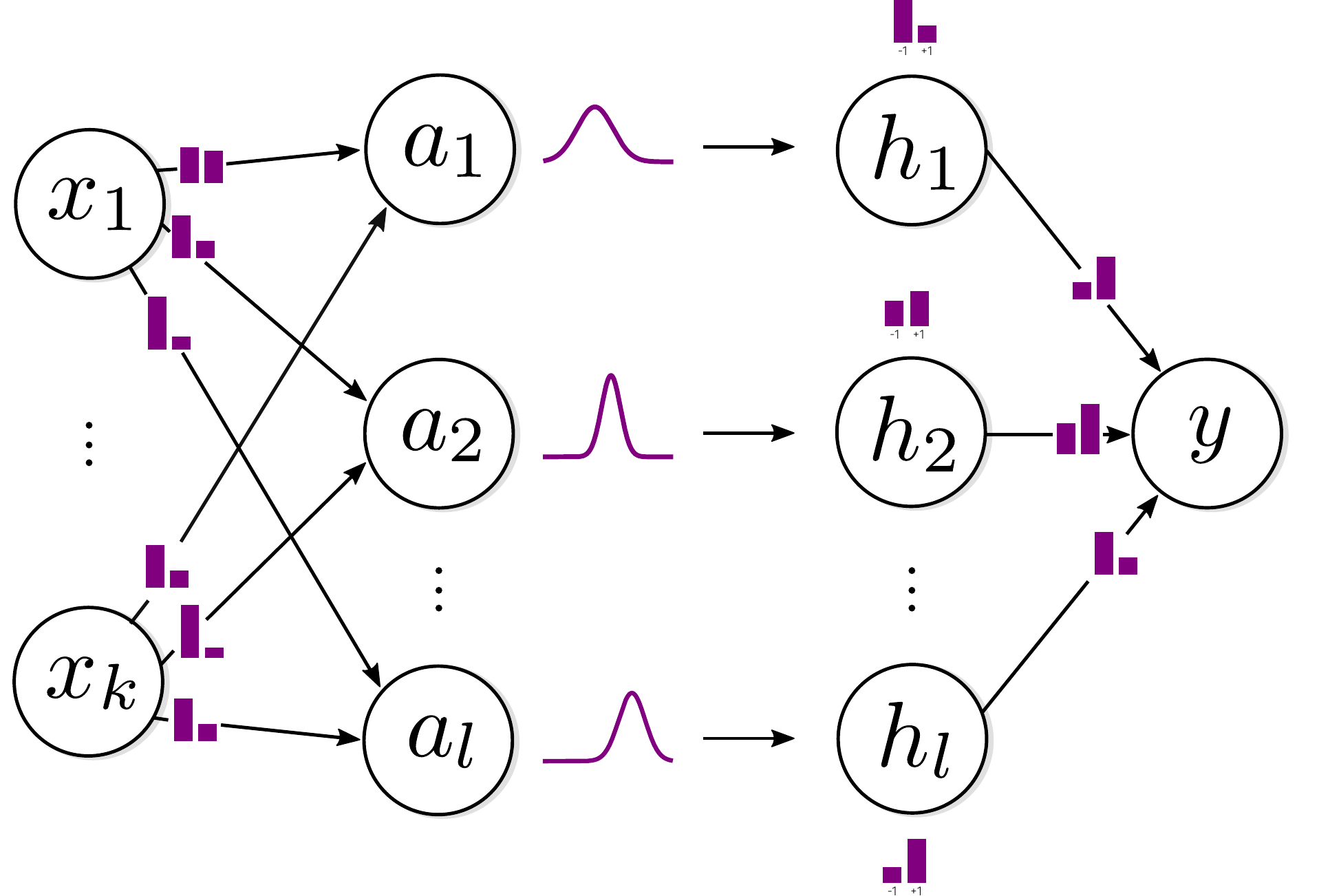}
  \caption{Graphical overview of a BLRNet layer. Given an input
    vector $\mbx$ and discrete distributions of the weights, a
    distribution over pre-activations $\mba$ is obtained using the central
    limit theorem.  This distribution is subsequently transformed using a
    discretization (or binarization) function, after which a discrete
    distribution over activations $\mbh$ is obtained. Samples from this
    distribution are the final result from a single layer in a
    BLRNet.}
  \label{fig:blrnet}
\end{wrapfigure}
We introduce a binary local reparametrization network using both
binary weights and binary activations.
Even when using binary weights and binary inputs to a layer, the
pre-activations can take on other values.  Often, an activation
function with a limited discrete co-domain -- such as
$\textrm{sign}(\cdot)$ -- is applied to the pre-activations to
constrain the activations of the network to some set of discrete
values. Unfortunately, when using this, one must deal with a
non-differentiable non-linear activation function.
Our method is based on the observation that when these activation
functions are applied to a normally distributed random pre-activation,
the computation involves one or more evaluations of the cumulative
density function (cdf) of the normal distribution, which is
differentiable. A binary (or discrete) sample can then be obtained
using the Concrete continuous relaxation of a discrete distribution.
Although, this leads to biased (but low variance) gradients, it can be
used to effectively optimize a network with discrete
nodes~\citep{maddison2016concrete}.

We extend the stochastic method for training binary weight networks
of~\citet{shayar2017learning} to allow for binary activations.  We
assume a Bernoulli distribution over $\{-1, +1\}$ for each parameter
in the network and leverage the (Lyapunov) central limit theorem to
obtain a normal distribution over the pre-activations in each layer.
Subsequently, a binarization activation function is applied to these
distributions in order to obtain a binary distribution over
activations. We call a network using these methods a Binary Local
Reparameterization Network, or BLRNet. A graphical overview of a
BLRNet layer is given in Figure~\ref{fig:blrnet}.

A consequence of applying the activation function to a random variable
is that any operation normally applied between the linear operation
and the activation function must also be applied to a random variable.
For this reason, we introduce an interpretation of Batch
Normalization~\citep{ioffe2015batch} and max pooling that can be
trained in a stochastic setting and applied in a deterministic
setting. Pseudo code for the full forward pass of a single layer,
including batch normalization and max pooling, is given in
Algorithm~\ref{algo:binlayer}.

\subsection{Stochastic Binary Activation}
\begin{figure}[h]
  \centering
  \includegraphics[width=0.8\textwidth]{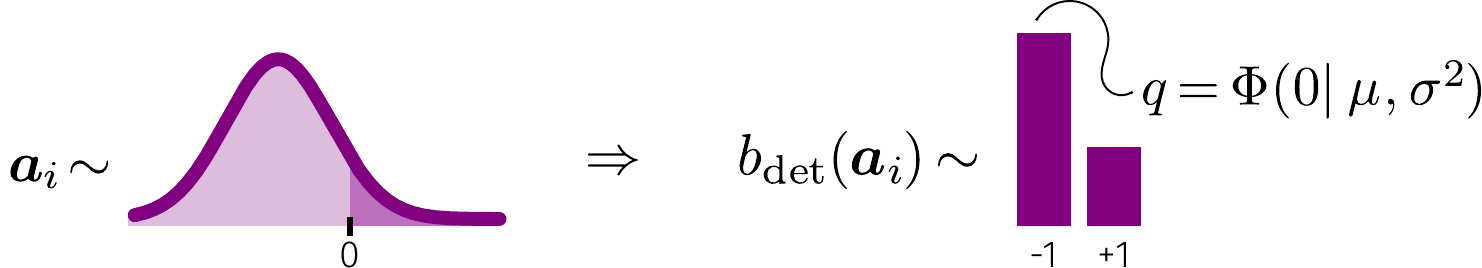}
  \caption{Given a random variable and a deterministic binarization
    function, the probability associated with each \emph{bin} of the
discrete output distribution is computed using the cdf of the
distribution of the input variable.  Although a deterministic
binarization function is used, a stochastic activation is obtained.}
  \label{fig:stochact}
\end{figure}

Since the output of a linear operation using binary inputs is not
restricted to be binary, it is required to apply a binarization
(non-linear) operation to the pre-activation in order to obtain binary
activations. Various works -- e.g., \citet{hubara2016binarized} and
\citet{rastegari2016xnor} -- use either deterministic or stochastic
binarization functions, i.e.,
\begin{align}
  b_{\text{det}}(a) = \begin{cases} +1 & \text{if $a \ge 0$} \\ -1 & \text{otherwise} \end{cases}\quad
  b_{\text{stoch}}(a) = \begin{cases} +1 & \text{with probability $p = \textrm{sigmoid}(a)$} \\ -1 & \text{with probability $1-p$} \end{cases}.
\end{align}
However, in the present case there is no such distinction since the
pre-activations are random variables: applying a deterministic
binarization function to a random pre-activation results in a
stochastic binary activation. Specifically, let $\mba_i \sim
\mathcal{N}(\mbu_i, \mbsigma^2_i)$ be a random activation obtained
using the CLT, then
\begin{align}
  \mba^s_i = b_{\text{det}}(\mba_i) &\sim \textrm{Bern}_{\pm}(q),\quad q = \Phi(0|\mu_i, \sigma^2_i), \\
  \intertext{where $\Phi(\cdot|\mbmu, \mbsigma^2)$ denotes the cdf of $\mathcal{N}(\mbmu, \mbsigma^2)$, and
$\Bernpm(q)$ denotes a Bernoulli distribution on $\{-1, 1\}$, such
that}
  P(\mba^s_i = -1) = q &,\ \ \text{and}\ \  P(\mba^s_i = 1) = 1-q.
\end{align}
During training, samples are drawn using the Concrete relaxation
method~\citep{maddison2016concrete}. By following these steps, both
the variance and the magnitude of the pre-activation are taken into
account when constructing the binary activation distribution, whereas
the stochastic activation function $b_{\text{stoch}}(a)$ only takes the
magnitude into account.  See Figure~\ref{fig:stochact} for a graphical
depiction of the stochastic binary activation.

At test time, a single binary weight instantiation $\hat{\mbB} \sim
p(\mbB)$ is obtained from the weight distribution and used to compute
the linear operation in a BLRNet layer. Subsequently,
$b_\text{det}(\cdot)$ is applied as non-linear activation. Hence, at
test time, a fully deterministic Binary Neural Network is obtained.

\begin{wrapfigure}[31]{r}{7.35cm}
\vspace{-1.5cm}
\begin{algorithm}[H]
  \KwIn{$\mba_{l-1}$, $\mbB \sim p(\mbB)$, $\tau$, $f(\cdot,
    \cdot)$, $\epsilon$, $\gamma$, $\beta$}
 \KwResult{Binary activation $\mba_l$}
 \vspace{5px}
 \tcp{CLT approximation}
 $\mbmu = f(\mathbb{E}[\mbB], \mba_{l-1})$\;
 $\mbsigma^2 = f(\mathbb{V}[\mbB], \mba_{l-1}^2)$\;
 \vspace{5px}
 \tcp{Batch normalization}
 $\mbm = \textrm{channel-wise-mean}(\mbmu)$\;
 $\mbv = \textrm{channel-wise-variance}(\mbmu, \mbsigma^2, \mbm)$\;
 $\mbmu = \gamma(\mbmu - \mbm)/\sqrt{\mbv + \epsilon} + \beta$\;
 $\mbsigma^2 = \gamma^2\mbsigma^2/(\mbv + \epsilon)$\;
 \vspace{5px}
 \tcp{Max pooling}
 \If{max pooling required}{
 $\mbn \sim \mathcal{N}(0, \mbI)$\;
 $\mbs = \mbmu + \mbsigma \odot \mbn$\;
 $\mbiota = \textrm{max-pooling-indices}(\mbs)$\;
 $\mbmu, \mbsigma^2 = \textrm{select-variable-at-indices}(\mbmu,
 \mbsigma^2, \mbiota)$\;}
 \vspace{5px}
 \tcp{Binarization and sampling}
 $\mbp \leftarrow \Phi(0|\mbmu, \mbsigma^2)$\;
 $\mba_l \sim \textrm{BinaryConcrete}(1-\mbp, \tau)$\;
 \Return{$\mba_l$}
 \vspace{5px}
 \caption{Pseudo code for forward pass of single layer in
   BLRNet. $\mba_{l-1}$ denotes the activation of the previous layer,
   $\mbB$ the random binary weight matrix, $\tau$ is the temperature
   used for the concrete distribution, $f(\cdot, \cdot)$ the linear
   transformation used in the layer, $\epsilon > 0$ a small
   constant for numerical stability, and $\gamma$ \& $\beta$ are the
   parameters for batch normalization.}
 \label{algo:binlayer}
\end{algorithm}
\end{wrapfigure}

\subsection{Batch Normalization and Pooling}
\label{sec:poolbn}
Other than a linear operation and an (non-linear) activation function,
Batch Normalization~\citep{ioffe2015batch} and pooling are two popular
building blocks for Convolutional Neural Networks.
For Binary Neural Networks, applying Batch Normalization to a
binarized activation will result in a non-binary result. Moreover, the
application of max pooling on a binary activation will result in a
feature map containing mostly $+1$s.  Hence, both operations must be
applied before binarization. However, in the BLRNet, the binarization
operation is applied before sampling. As a consequence, the Batch
Normalization and pooling operations can only be applied on random
pre-activations. For this reason, we define these methods for random
variables. Although there are various ways to define these operation
in a stochastic fashion, our guiding principle is to only leverage
stochasticity during training, i.e., at test time, the stochastic
operations are replaced by their conventional implementations and
learned parameters learned in the stochastic setting must be transferred
to their deterministic counterparts.

\subsubsection{Stochastic Batch Normalization}
Batch Normalization (BN)~\citep{ioffe2015batch} --- including an affine transformation ---
is defined as follows:
\begin{align}
  \hat{\mba}_i = \frac{\mba_i - \mbm}{\sqrt{\mbv + \epsilon}}\gamma + \beta,
  \label{eq:batchnorm}
\end{align}
where $\mba_i$ denotes the pre-activation before BN, $\hat{\mba}$ the
pre-activation after BN, and $\mbm$ \& $\mbv$ denote the sample mean
and variance of $\{\mba_i\}_{i=1}^M$, for an $M$-dimensional
pre-activation, respectively. In essence, BN translates and scales
the pre-activations such that they are approximately zero mean and
have unit variance, followed by an affine transformation. Hence, in
the stochastic case, our aim is that samples from the pre-activation
distribution after BN also have approximately zero mean and
unit variance---to ensure that the stochastic batch normalization can be transfered to a deterministic binary neural network. This is achieved by subtracting the
population mean from each pre-activation random variable and by
dividing by the population variance. 
However, since $\mba_i$ is a random variable in the BLRNet, simply using
the population mean and variance equations will result in a non-standard output.
Instead, to ensure a standard distribution over activations, we compute the expected population mean
and variance under the pre-activation distribution:
\begin{align}
  \mathbb{E}_{p(\mba|\mbB, \mbh)}[\mbm]
  &= \mathbb{E}\left[\frac{1}{M}\sum_{i=1}^M \mba_i\right]
  = \frac{1}{M}\sum_{i=1}^M \mathbb{E}\left[\mba_i\right]
    \label{eq:expmean}
  = \frac{1}{M}\sum_{i=1}^M \mbmu_i \\
  \mathbb{E}_{p(\mba|\mbB, \mbh)}[\mbv]
  &= \mathbb{E}\left[ \frac{1}{M-1} \sum_{i=1}^M (\mba_i - \mathbb{E}[\mbm])^2\right] 
  = \frac{1}{M-1} \left\{ \sum_{i=1}^K \mbsigma^2_i + \sum_{i=1}^M \left( \mbmu_i - \mathbb{E}[\mbm]\right)^2\right\},
    \label{eq:expvar}
\end{align}
where $M$ is the total number of activations and $\mba_i \sim
\mathcal{N}(\mbmu_i, \mbsigma_i)$ are the random pre-activations.  By
substituting $\mbm$ and $\mbv$ in Equation~\ref{eq:batchnorm} by
Equation~\ref{eq:expmean} and~\ref{eq:expvar}, we obtain
the following batch normalized Gaussian distributions for the
pre-activations:
\begin{align}
  \hat{\mba}_i = \frac{\mba_i - \mathbb{E}[\mbm]}{\sqrt{\mathbb{E}[\mbv] + \epsilon}}\gamma + \beta \quad \Rightarrow \quad
\hat{\mba}_i \sim
  \mathcal{N}\left(
  \frac{\mbmu_i - \mathbb{E}[\mbm]}{\sqrt{\mathbb{E}[\mbv] + \epsilon}}\gamma + \beta,
  \frac{\gamma^2}{\mathbb{E}[\mbv] + \epsilon}\mbsigma^2_i\right).
\end{align}
Note that this assumes a single channel, but is easily extended to 2d
batch norm in a similar fashion as conventional Batch Normalization. 

\subsubsection{Stochastic Max Pooling}
In the deterministic case, pooling applies an aggregation operation to
a set of (spatially oriented) pre-activations. Here we discuss max
pooling for stochastic pre-activations, however, similar considerations
apply for other types of aggregation functions.

\begin{figure}[h]
  \centering
  \includegraphics[width=\textwidth]{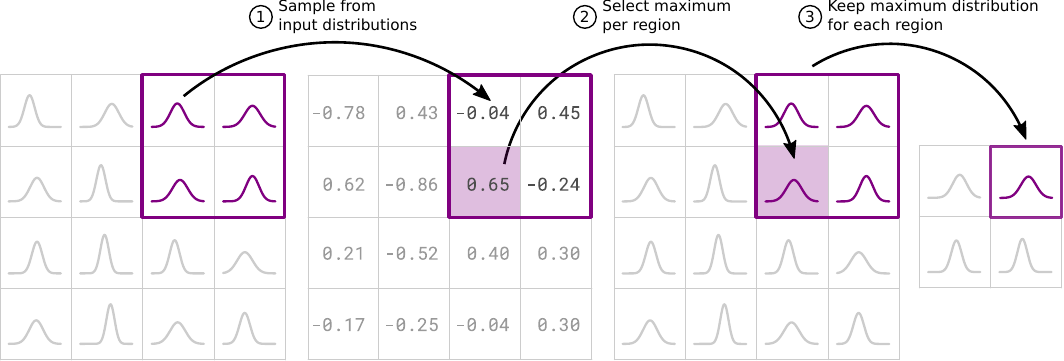}
  \caption{Max pooling for random variables is performed by taking a single sample from
    each of the input distributions. The output random variable for
    each pooling region is the random variable that is associated with the
    maximum sample.}
  \label{fig:stochmaxpool}
\end{figure}

In the case of max-pooling, given a spatial region containing
stochastic pre-activations $\mba_1, \ldots, \mba_K$, we aim to
stochastically select one of the $\mba_i$. Note that, although the
distribution of $\max(\mba_1, \ldots, \mba_K)$ is
well-defined~\citep{nadarajah2008exact}, it's distribution is not
Gaussian and thus does not match one of the input distributions.
Instead, we sample one of the input random variables in every
spatial region according to the probability of that variable being 
greater than all other variables, i.e., $\rho_i = p(\mba_i >
\mbz_{\textbackslash i})$, where $\mbz_{\textbackslash i} =
\max(\{\mba_j\}_{j\neq i})$.
$\rho_i$ could be obtained by evaluating the CDF of
$(\mbz_{\textbackslash i} - \mba_i)$ at 0, but to our knowledge this has
no analytical form. Alternatively, we can use monte-carlo integration
to obtain $\mbrho$:
\begin{align}
  \mbrho \approx \frac{1}{L} \sum_{l=1}^L \textrm{one-hot}(\arg\max \mbs^{(l)}),\quad \mbs^{(l)} \sim p(\mba_1, \mba_2, \ldots, \mba_K) = \prod_{i=1}^K \mathcal{N}(\mbmu_i, \mbsigma_i^2)
\end{align}
where $\textrm{one-hot}(i)$ returns a $K$-dimensional one-hot vector
with the $i$th elements set to one. The pooling index $\mbiota$ is
then sampled from $\textrm{Cat}(\mbrho)$.  However, more efficiently,
we can sample $\mbs \sim p(\mba_1, \ldots, \mba_K)$ and select the
index of the maximum in $\mbs$, which is equivalent sampling from
$\textrm{Cat}(\mbrho)$. Hence, for a given max pooling region, it is
sufficient to obtain a single sample from each normal distribution
associated with each pre-activation and \emph{keep} the random
variable for which this sample is maximum. A graphical overview of
this is given in Figure~\ref{fig:stochmaxpool}.

Other forms of stochastic or probabilistic max pooling were introduced by \cite{lee2009convolutional} and \cite{zeiler2013stochastic}, however, in both cases
a single activation is sampled based on the magnitude of the activations. In
contrast, in our procedure we stochastically propagate one of the input distributions over activations.

\subsection{Weight Initialization}
\label{sec:weightinit}
The weights for a BLRNet are initialized using a pre-trained full
precision network with the same architecture. This initializes the
convolutional filters with more structure than a random
initialization. This is desirable as in order to flip the value of a
weight, the parameter governing the weight has to pass through a high
variance regime, which can slow down convergence considerably.

We use the weight transfer method introduced
by~\cite{shayar2017learning} in which the parameters of the weight
distribution for each layer are initialized such that the expected
value of the random weights equals the full precision weight divided
by the standard deviation of the weights in the given layer. Since not
all rescaled weights lay in the $[-1, 1]$ range, all weight probabilities 
are clipped between $[0.05, 0.95]$. This transfer method transfers the
structure present in the filters of the full precision network and
ensures that a significant part of the parameter distributions is
initialized with low variance. 

\subsection{Deterministic Binary Neural Network}
\label{sec:detnet}
During training, a stochastic network is optimized. However, on
hardware one wants to leverage all the advantages of a fully binary
neural network. Therefore, we obtain one or multiple binary neural
networks from the parameter distribution $p(\mbB)$ at test time. We
consider two options: the MAP, or most likely, estimate denoted {\sc
BLRNet-map}, and an ensemble consisting of 2, 5, or 16 samples from
the parameter distribution denoted {\sc BLRNet-$x$}. Note that, even
when using multiple binary neural networks in an ensemble, the
ensemble is still more efficient in terms of computation and memory
when compared to a full precision alternative. The ensemble
predictions are obtained by summing the $\log$ softmax probability for
each member of the ensemble and selecting the class with the maximum
resulting value.

Since the trained weight distribution is not fully deterministic, the
sampling of individual weight instantiations will result in a shift of
the batch statistics. As a consequence, the learned batch norm
statistics no longer closely match the true statistics. This is
alleviated by re-estimating the batch norm statistics based on (a
subset of) the training set after weight sampling using a moving mean
and variance estimator.  We observed competitive results using as
little as 5 batches from the training set. However, given the iid
nature of the datasets considered in this work, these could be
estimated more efficiently by directly computing the batch statistics
using a smaller sample.

\subsection{Bayesian Interpretation and Considerations}
The BLRNet can be interpreted as a Bayesian neural network, i.e.,
$p(\mbB)$ is a variational approximation to the true posterior
$p(\mbB|\mbY, \mbX)$, where $\mbX$ denotes the training inputs and
$\mbY$ the training targets. In that case, assuming a uniform prior
distribution on the binary weights, it can be optimized by maximizing
the following variational lower bound:
\begin{align}
  \mathcal{L}(\mbtheta) =
  \mathbb{E}_{p(\mbB)}[\log p(\mbY|\mbX,\mbB)] + \mathbb{H}[p(\mbB)].
\end{align}
This objective favors models with both high accuracy and uncertainty
on the weights. This allows one to estimate prediction uncertainty
originating from the model uncertainty. However, in the present case,
we aim to obtain a single best predictive model. Therefore we deviate
from the strict approximate Bayesian training and use the following
objective:
\begin{align}
  \hat{\mathcal{L}}(\mbtheta) = \mathbb{E}_{p(\mbB)}[\log p(\mbY|\mbX,\mbB)]
  - \underbrace{\beta||\sigma(\mbW)\odot(1 - \sigma(\mbW))||_1}_{\text{variance regularizer}},
\end{align}
where $\sigma(\mbW)$ contains the probabilities for the binary weight
distributions.  In contrast to the variational objective, this
objective favors model with high accuracy and low uncertainty on the
weight distributions. This regularizer is proportional to the
variance of the weight distribution and therefore we refer to it as
the variance regularizer. In~\cite{shayar2017learning} it is also used
for the binary weight network and is called the beta parameter.

\section{Experiments}
\label{sec:experiments}
We evaluated the BLRNet on the MNIST and CIFAR-10 benchmarks and
compare the results to Binarized Neural
Networks~\citep{hubara2016binarized}, since the architecture of the
deterministic networks obtained by training the BLRNet are equivalent.

\subsection{Experimental Details}
All BLRNetworks are trained using cross-entropy loss plus a weight
decay term scaled by $10^{-4}$ on the parameters of the final softmax
layer and a variance regularizer on the parameters of the weight
distributions rescaled by $\beta = 10^{-6}$. Note that this training objective
deviates from the variational lower bound as we aim to optimize the
BLRNetwork to obtain a single best deterministic network, instead of
obtaining a posterior that captures model uncertainty. The weights
for all networks are initialized using the transfer method described
in Section~\ref{sec:weightinit}. All models are optimized using
Adam~\citep{kingma2014adam} with an initial learning rate of
$10^{-2}$, a batch size of 128, and a validation loss plateau learning
rate decay scheme. We keep the temperature for the binary concrete
distribution static at $1.0$ during training. All models are
implemented using PyTorch~\citep{paszke2017automatic}. All models are
optimized until convergence, after which the best model is selected
based on a validation set.

For Binarized Neural Networks we use the training procedure described
by~\cite{hubara2016binarized}, i.e., a squared hinge loss and layer
specific learning rates that are determined based on the Glorot
initialization method~\citep{glorot2010understanding}.

\subsection{MNIST}
The MNIST dataset consists of of 60K training and 10K test
28$\times$28 grayscale handwritten digit images, divided over 10
classes. The images are pre-processed by subtracting the global pixel
mean and dividing by the global pixel standard deviation. No other
form of pre-processing or data augmentation is used. For MNIST,
we use the following architecture:
\begin{align*}
  32\textrm{C}3 \ - \ 
  \textrm{MP2} \ - \ 
  64\textrm{C}3 \ - \
  \textrm{MP2} \ - \ 
  512\textrm{FC} \ - \ 
  \textrm{SM}10
\end{align*}
where $X\textrm{C}3$ denotes a binary convolutional layer using $3
\times 3$ filters and $X$ output channels, followed by (stochastic)
batch normalization and binarization of the activations,
$Y\textrm{FC}$ denotes a fully connected layer with $Y$ output
neurons, $\textrm{SM}10$ denotes a softmax layer with 10 outputs, and
$\textrm{MP}2$ denotes $2\times2$ (stochastic) max pooling with stride
2. Note that if a convolutional layer is followed by a max pooling
layer, the binarization is only performed after max pooling. Results
are reported in
Table~\ref{table:results}.

\subsection{CIFAR-10}
The CIFAR-10~\citep{krizhevsky2009learning} dataset consists of 50K
training and 10K test $32\times 32$ RGB images divided over 10
classes. The last 5,000 images from the training set are used as
validation set. We perform two different experiments using
CIFAR-10. In the first the images are only pre-processed by
subtracting the channel-wise mean and dividing by the standard
deviation. In the second experiment we perform ZCA-whitening
on the images.  For both, the same architecture
as~\cite{shayar2017learning} is used, i.e.,
\begin{align*}
  2 \times 128\textrm{C}3 \ - \ 
  \textrm{MP2} \ - \ 
  2 \times 256\textrm{C}3 \ - \
  \textrm{MP2} \ - \ 
  2 \times 512\textrm{C}3 \ - \
  \textrm{MP2} \ - \ 
  1024\textrm{FC} \ - \ 
  \textrm{SM}10
\end{align*}
where we use the same notation as in the previous section.  The
Binarized Neural Network baseline uses the same architecture, except
for one extra 1024 neuron fully connected layer.  During training, the
training set is augmented using random 0px to 4px translations and random
horizontal flips.  Results for both experiments -- and ensembles --
are reported in Table~\ref{table:results}.

\begin{table}
  \begin{center}
    \caption{Test accuracy on MNIST and CIFAR-10 for Binarized
NN~\citep{hubara2016binarized}, BLRNet, and a full precission network
(FPNet). BLRNet-map refers to a deterministic BLRNet using the map
estimate, and BLRNet-$X$ refers to an ensemble of $X$ networks, each
sampled from the same weight distribution. For the ensemble results
both mean and standard deviation are presented obtained from sampling
multiple ensembles from the weight distribution.}
  \label{table:results}
  \begin{tabular}{l c c c} 
    \toprule
    & \sc{mnist} & \sc{cifar-10} & \sc{cifar-10 (white)} \\
    \midrule
    \sc{Binarized NN} & 99.17\% & 88.17\% & 88.56\%  \\
    {\sc BLRNet-map}      & 99.00\% & 88.61\% & $88.96\%$  \\
    \midrule
    {\sc BLRNet-2}    & $99.09\pm0.05\%$ & $89.51\pm0.25\%$ & $89.78\pm0.16\%$ \\
    {\sc BLRNet-5}    & $99.13\pm0.03\%$ & $90.66\pm0.12\%$ & $90.48\pm0.13\%$ \\
    {\sc BLRNet-16}   & $99.15\pm0.03\%$ & $91.22\pm0.08\%$ & $90.82\pm0.08\%$ \\
    \midrule
    \sc{FPNet}        & 99.48\% & 92.36\% & 92.45\%  \\
    \bottomrule
   \end{tabular}
   \end{center}
\end{table}

\subsection{Effect of Batch Statistics Re-estimation}
As discussed in Section~\ref{sec:detnet}, after sampling the
parameters of a deterministic network the batch statistics used by
Batch Normalization must be re-estimated. Figure~\ref{fig:bnre} shows
the results obtained using a various number of batches from the
training set to re-estimate the statistics. This shows that even a
small number of samples is sufficient to estimate the statistics.

\subsection{Ensemble Based Uncertainty Estimation}
\begin{figure}
\vspace{-.6cm}
  \centering
  \begin{subfigure}[t]{0.48\textwidth}
  \includegraphics[width=\textwidth]{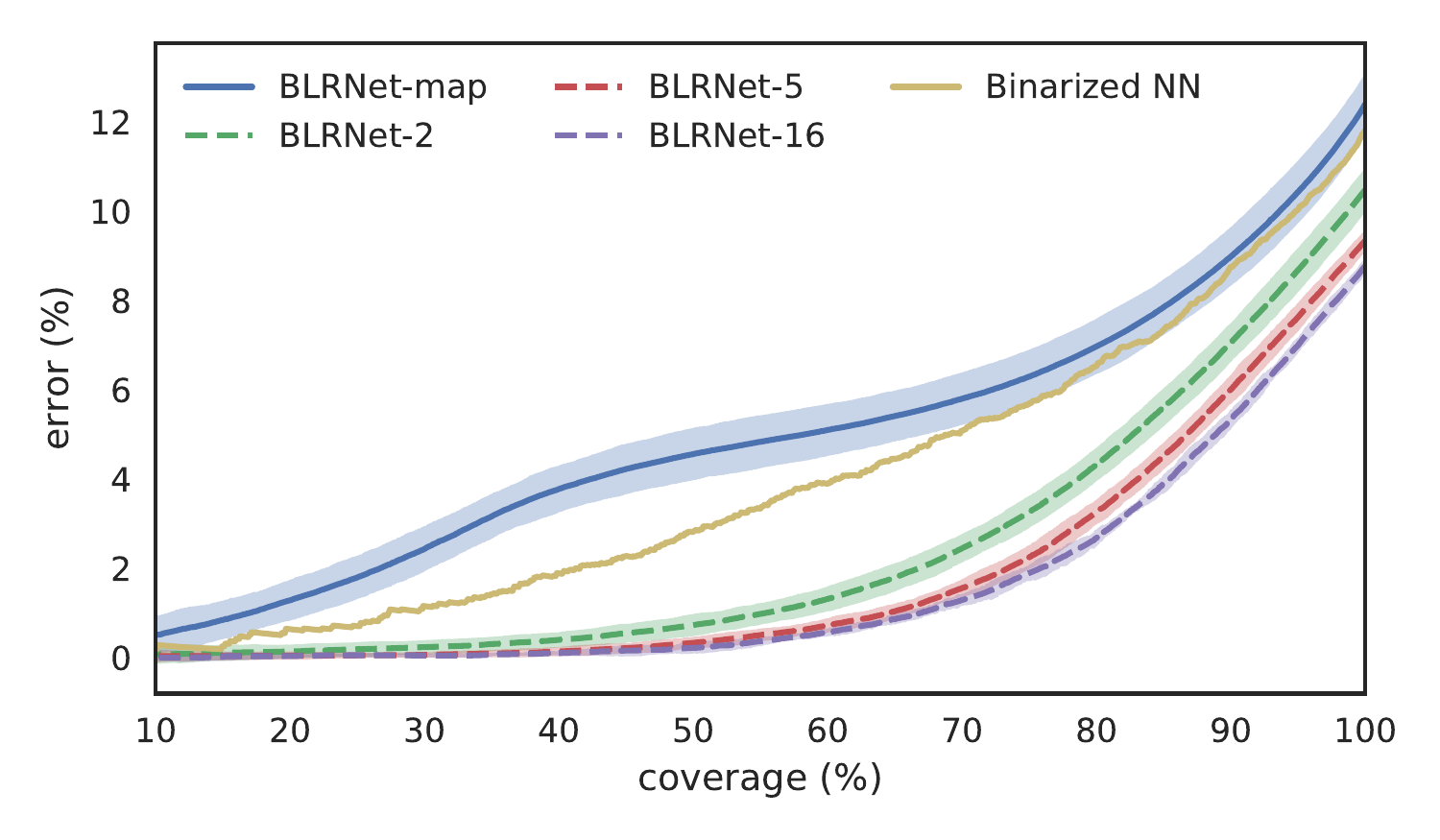}
  \caption{Error coverage curve for CIFAR-10.}
  \label{fig:errcov}
  \end{subfigure}
  \begin{subfigure}[t]{0.48\textwidth}
  \includegraphics[width=\textwidth]{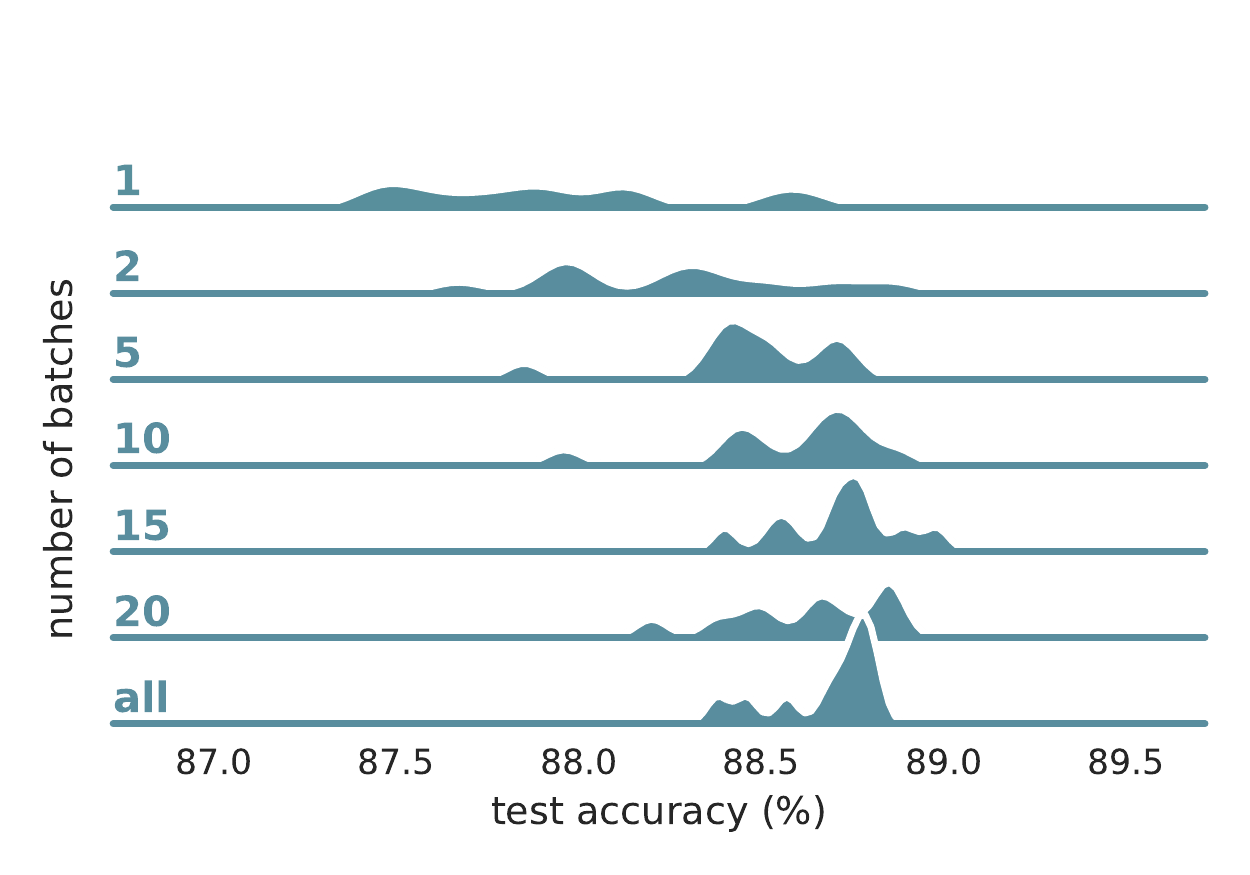}
  \caption{Test set performance with and without BN re-estimation on CIFAR-10.}
  \label{fig:bnre}
  \end{subfigure}
  \caption{Error coverage curve and batch statistic re-estimation results
    for CIFAR-10.}
  \label{fig:riscov}
\end{figure}

As presented in Table~\ref{table:results} the accuracy improves when
using an ensemble. Moreover, the predictions of the ensemble members
can be used to obtain an estimate of the certainty of the ensemble as
a whole. To evaluate this, we plot an error-coverage curve~\citep{geifman2017selective} in
Figure~\ref{fig:errcov}. This curve is obtained by sorting the samples
according to a statistic and computing the error percentage in the top
$x$\% of the samples -- according to the statistic. For the Binarized
Neural Network and BLRNet-MAP the highest softmax score is used,
whereas for the ensembles the variance in the prediction of the top
class.  The figure suggests that the ensemble variance is a better
estimator of network certainty, and moreover, the estimation improves
as the ensemble sizes increases.

\subsection{Abblation studies}
\begin{figure}
    \centering
    \includegraphics[width=\textwidth]{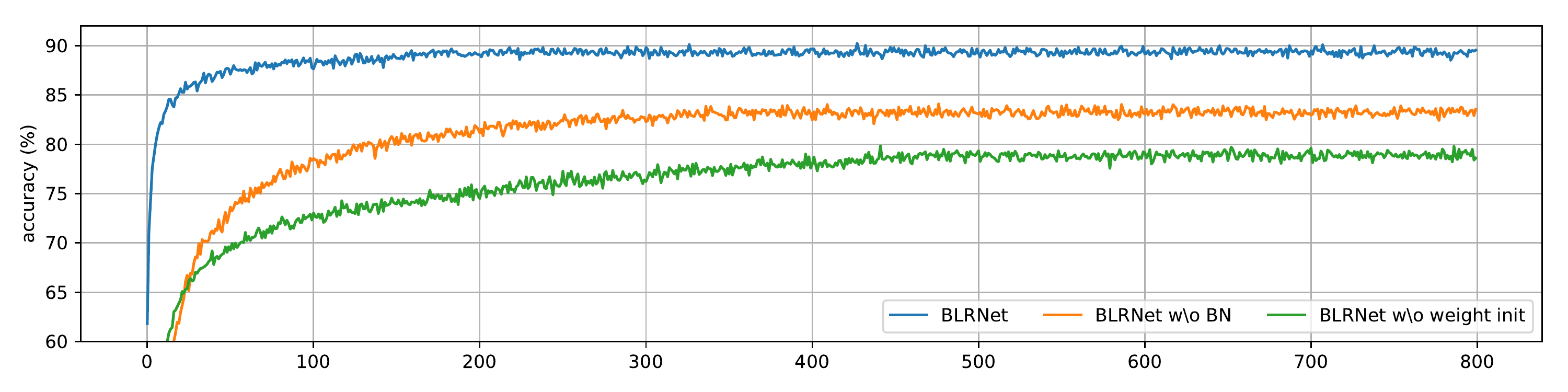}
    \caption{Accuracy on validation set during training, i.e., using stochastic weights,
    local reparametrization and binary activation sampling.}
    \label{fig:abl}
\end{figure}

We perform an abblation studie on both the use of (stochastic) batch normalization and
the use of weight transfer for the BLRNet on CIFAR-10. For batch normalization, we removed
all batch normalization layers from the BLRNet and retrained the BLRNet on CIFAR-10. This
resulted in a test set accuracy of 79.24\%. For the weight initialization experiment,
the BLRNet weights are initialized using the Xavier initialization scheme~\cite{glorot2010understanding} and was trained on CIFAR-10. When using Xavier initialization, a test set accuracy of 75.07\% was obtained. These results
are also presented in Table~\ref{table:abbres}. Moreover, the accuracy on the validation set during
training is presented in Figure~\ref{fig:abl}. Note that these numbers are obtained \emph{without}
sampling a binarized network from the weight distribution, i.e., local reparametrization and binary activation
samples are used. The BLRNet that used both weight transfer and stochastic
batch normalization results in a significant performance improvement, indicating that both stochastic batchnorm and weight transfer are necessary components for the BLRNet.

\begin{table}
\begin{center}
  \caption{Abblation results}
  \label{table:abbres}
  \begin{tabular}{l c c c} 
    \toprule
      & \sc{BLRNet} & \sc{BLRnet} without BatchNorm & \sc{BLRNet} Xavier init \\
    \midrule
    \sc{CIFAR-10} & 88.61\% & 79.24\% & 75.07\% \\
    \bottomrule
   \end{tabular}
   \end{center}
\end{table}

\section{Conclusion}
We have presented a stochastic method for training Binary Neural
Networks. The method is evaluated on multiple standardized benchmarks
and reached competitive results. The BLRNetwork has various
advantageous properties as a result of the training method. The weight
distribution allows one to generate ensembles online which results in
improved accuracy and better uncertainty estimations. Moreover, the
Bayesian formulation of the BLRNetwork allows for further pruning of
the network, which we leave as future work.

\bibliographystyle{apalike}
\bibliography{ref.bib}

\end{document}